\documentclass[runningheads]{llncs}
\usepackage[T1]{fontenc}
\usepackage{graphicx}
\usepackage{multirow}
\usepackage{amsmath}
\usepackage{hyperref}       
\usepackage{url}            
\usepackage{booktabs}       
\usepackage{amsfonts}       
\usepackage{nicefrac}       
\usepackage{microtype}      
\usepackage{xcolor}  
\usepackage{multirow}
\usepackage{authblk}
\begin{document}
\title{GTA-Net: Cooperative Game Theory for Vision-Language Alignment in Chest X-Ray Report Generation}
%
%
\author{Saif ur Rehman Khan\inst{1,2}\orcidID{0000-0002-0768-5239} \and
Imad Ahmed Waqar\inst{1}\orcidID{0000-0002-0728-5209}
Sebastian Vollmer\inst{2,3}\orcidID{0000-0002-7863-0462} \and
Andreas Dengel\inst{1,2,3}\orcidID{0000-0002-6100-8255}\and
Muhammad Nabeel Asim\inst{1,2,3,4}\orcidID{0000-0001-5507-198X}}
\authorrunning{Saif. Khan et al.}
%
\institute{Department of Computer Science, Rhineland-Palatinate Technical University of Kaiserslautern-Landau, Kaiserslautern 67663, Germany \and
German Research Center for Artificial Intelligence (DFKI), Kaiserslautern 67663, Germany \and
IntelligentX GmbH (intelligentx.com), Kaiserslautern, Germany \and
Department of Core Informatics, Graduate School of Informatics ,Osaka Metropolitan University, Saka, 599-8531, Japan \\
\email{\{saif\_ur\_rehman.khan,sebastian.vollmer,andreas.dengel,muhammad\_nabeel.asim\}@dfki.de} \and
\email{maadiworks@gmail.com}
}
\maketitle              
\begin{abstract}
Automated chest X-ray report generation requires precise cross-modal grounding to ensure clinically reliable descriptions. However, existing vision–language models rely on implicit attention mechanisms that fail to enforce explicit region–word correspondence and disease-level consistency. We propose Game-Theoretic Alignment Network (GTA-Net), a vision–language framework that formulates report generation as a cooperative game-theoretic alignment problem. The model introduces a BinaryGameAligner that models interactions between image regions and text tokens using similarity-based payoff matrices with Shapley-inspired importance weighting. To enforce clinical semantics, we further develop a Disease-Aware Ternary Aligner, which captures joint interactions among images, reports, and structured disease concepts. GTA-Net combines a Swin-based visual encoder with a LoRA-adapted large language model and is trained with a unified objective for generation and alignment. Experiments on CheXpertPlus and IU-XRay demonstrate state-of-the-art performance across standard generation metrics and improved clinical consistency, highlighting the effectiveness of explicit game-theoretic alignment for medical vision–language generation.
\begin{center}
  \url{https://anonymous.4open.science/r/R2-09A3/}.
\end{center}
\end{abstract}

\section{Introduction}
\label{sec:intro}
Automatic radiology report generation has emerged as a promising application of deep learning, with the potential to reduce reporting workload and improve diagnostic consistency~\cite{ref1}. Unlike standard computer vision tasks that focus on object recognition, this problem requires generating structured clinical narratives that capture the presence, absence, and uncertainty of multiple pathologies~\cite{ref2}. The rapidly increasing volume of medical imaging has created a critical bottleneck in clinical workflows, as report writing often requires more time than image acquisition itself~\cite{ref3}. Consequently, automated report generation has become an important research direction at the intersection of computer vision and medical imaging.
\begin{figure*}[t]
\centering
\includegraphics[width=0.99\textwidth]{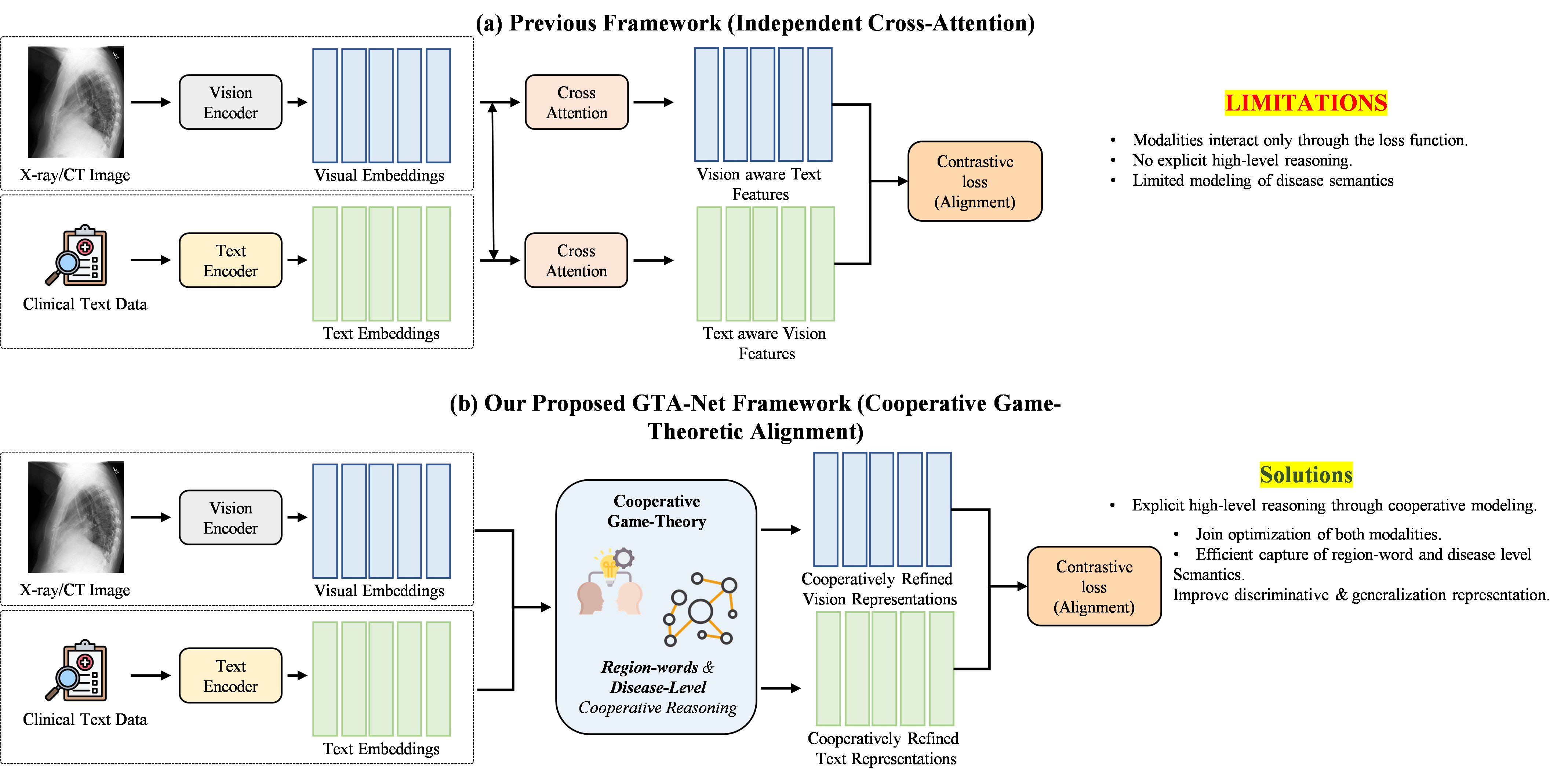}
\caption{Comparison between (a) Existing framework and (b) GTA-Net framework.}
\label{fig5}
\end{figure*}
Recent encoder–decoder approaches~\cite{ref4} unify multimodal acquisition, representation learning, fusion, and report generation, and are typically evaluated on public chest radiography benchmarks such as CheXpertPlus and IU-XRay. Despite these advances, cross-modal alignment remains largely implicit, driven by cross-entropy objectives and end-to-end optimization without explicit guarantees of semantic correspondence. While attention mechanisms improve fluency and local coherence, they do not provide reliable or theoretically grounded estimates of region-to-text contribution, limiting interpretability and hindering clinical trust.

Recent medical report generation methods increasingly rely on knowledge graphs to model anatomical and relational structure ~\cite{ref5}. However, these approaches ~\cite{ref6,ref7} depend on manually curated resources, making them costly to scale and susceptible to error propagation. In addition, data imbalance and the inherent complexity of clinical reports continue to limit both model performance and reliable evaluation. More fundamentally, existing models~\cite{ref6,ref7} exhibit a multi-granularity semantic gap, often producing repetitive, template-driven outputs while failing to capture rare but clinically significant findings. Previous work~\cite{ref5} shows that purely generative models struggle with such rare phenomena, motivating hybrid retrieval generation strategies to improve coverage and accuracy ~\cite{ref8}. Nevertheless, these approaches do not explicitly model hierarchical semantic dependencies, where fine-grained region–word alignment and coarse-grained disease-level consistency must be jointly optimized. While recent multi-grained and contrastive methods offer partial improvements, they lack a unified and principled framework for explicit cross-level semantic alignment without reliance on external supervision.

We reformulate vision–language alignment in radiology report generation as an explicit optimization objective, rather than an byproduct of sequence modeling. Drawing on cooperative game theory, we model visual regions, textual tokens, and disease concepts as interacting players whose joint contributions can be systematically quantified, enabling alignment to be learned without external supervision. Building on this formulation, we propose GTA-Net that captures both fine- and coarse-grained cross-modal interactions through cooperative alignment modules, jointly optimized with report generation in an end-to-end framework (Figure~\ref{fig5}).
\section{Methodology}
\label{sec:method}
\subsection{Problem Formulation}

Let $\mathcal{D} = \{(I_i, Y_i)\}_{i=1}^{N}$ denote a dataset of chest X-ray images $I_i$ and their corresponding radiology reports $Y_i = \{y_1, y_2, \dots, y_T\}$, where $y_t$ represents the $t$-th token in the report. The goal of automated radiology report generation is to learn a mapping function $f_{\theta}: I \rightarrow Y$ that generates a clinically accurate and coherent report conditioned on the input image.

A common approach ~\cite{ref5,ref6, ref7} is to use an encoder–decoder framework, where a visual encoder extracts image features $V = \{v_1, v_2, \dots, v_M\}$ from the input image, and a language decoder generates the report token-by-token:
\begin{equation}
P(Y|I) = \prod_{t=1}^{T} P(y_t \mid y_{<t}, V).
\end{equation}

However, in existing methods, the alignment between visual features and textual tokens is implicitly learned through attention mechanisms and maximum likelihood training, without explicit supervision or principled quantification of their interactions. This often leads to suboptimal ~\cite{ref8} semantic grounding, particularly across multiple levels of granularity.

To address this limitation, we reformulate vision language alignment as a cooperative game among multiple players. Specifically, we consider visual regions $V$, textual tokens $T$, and disease concepts $C$ as players in a game, where their interactions contribute to the overall alignment objective.

\paragraph{Binary Game Formulation}
At the fine-grained level, we model the interaction between visual features and textual tokens as a two-player cooperative game. A payoff function $\phi(v_i, y_j)$ is defined to measure the contribution of a visual region $v_i$ to a textual token $y_j$, typically computed via similarity in a shared embedding space:
\begin{equation}
\phi(v_i, y_j) = \text{sim}(v_i, e_j),
\end{equation}
where $e_j$ is the embedding of token $y_j$. The overall alignment is obtained by aggregating these contributions using a cooperative value function, enabling the model to estimate the importance of each region–token pair.

\paragraph{Ternary Game Formulation.}
To incorporate high-level semantic reasoning, we extend the formulation to a three-player cooperative game involving visual features $V$, textual tokens $T$, and disease concepts $C = \{c_1, c_2, \dots, c_K\}$. The joint payoff function is defined as:
\begin{equation}
\phi(v_i, y_j, c_k) = \text{sim}(v_i, c_k) + \text{sim}(e_j, c_k),
\end{equation}
which captures both image–disease and text–disease compatibility. This formulation enables the model to align visual and textual modalities within a shared clinical concept space.

\paragraph{Learning Objective.}
The overall training objective combines the report generation loss with the proposed game-theoretic alignment objectives:
\begin{equation}
\mathcal{L} = \mathcal{L}_{gen} + \lambda_1 \mathcal{L}_{binary} + \lambda_2 \mathcal{L}_{ternary},
\end{equation}
where $\mathcal{L}_{gen}$ is the standard cross-entropy loss for report generation, and $\mathcal{L}_{binary}$ and $\mathcal{L}_{ternary}$ correspond to alignment objectives derived from the respective game formulations.

This formulation enables explicit modeling of cross-modal interactions at multiple semantic levels, providing a principled framework for improving both alignment and interpretability in radiology report generation.

The contributions of this paper can be summarized as follows:
\begin{itemize}

\item Game-Theoretic Alignment for Vision Language Modeling:
We introduce GTA-Net, a novel framework that formulates cross-modal alignment in radiology report generation as a cooperative game. By modeling visual regions, textual tokens, and disease concepts as interacting plays, our approach provides an explicit and principled mechanism for measuring and optimizing semantic correspondence without requiring additional annotations.

\item Multi-Level Cooperative Alignment:
We design two complementary game-theoretic modules to capture alignment at different semantic granularities. A binary game aligner models fine-grained region–word interactions, while a disease-aware ternary aligner incorporates high-level clinical concepts to ensure global diagnostic consistency.

\item Implicit Supervision via Game Dynamics:
The proposed alignment mechanisms act as auxiliary objectives that guide the learning process, enabling the model to jointly optimize report generation and semantic grounding without explicit region-level supervision.

\item Improved Interpretability and Clinical Relevance:
By quantifying feature contributions through game-theoretic payoffs, GTA-Net enhances interpretability and produces clinically coherent reports, addressing key limitations of existing implicit alignment approaches.

\end{itemize}

\subsection{Benchmark Dataset}
\begin{figure}[h]
\centering

\includegraphics[width=0.6\textwidth]{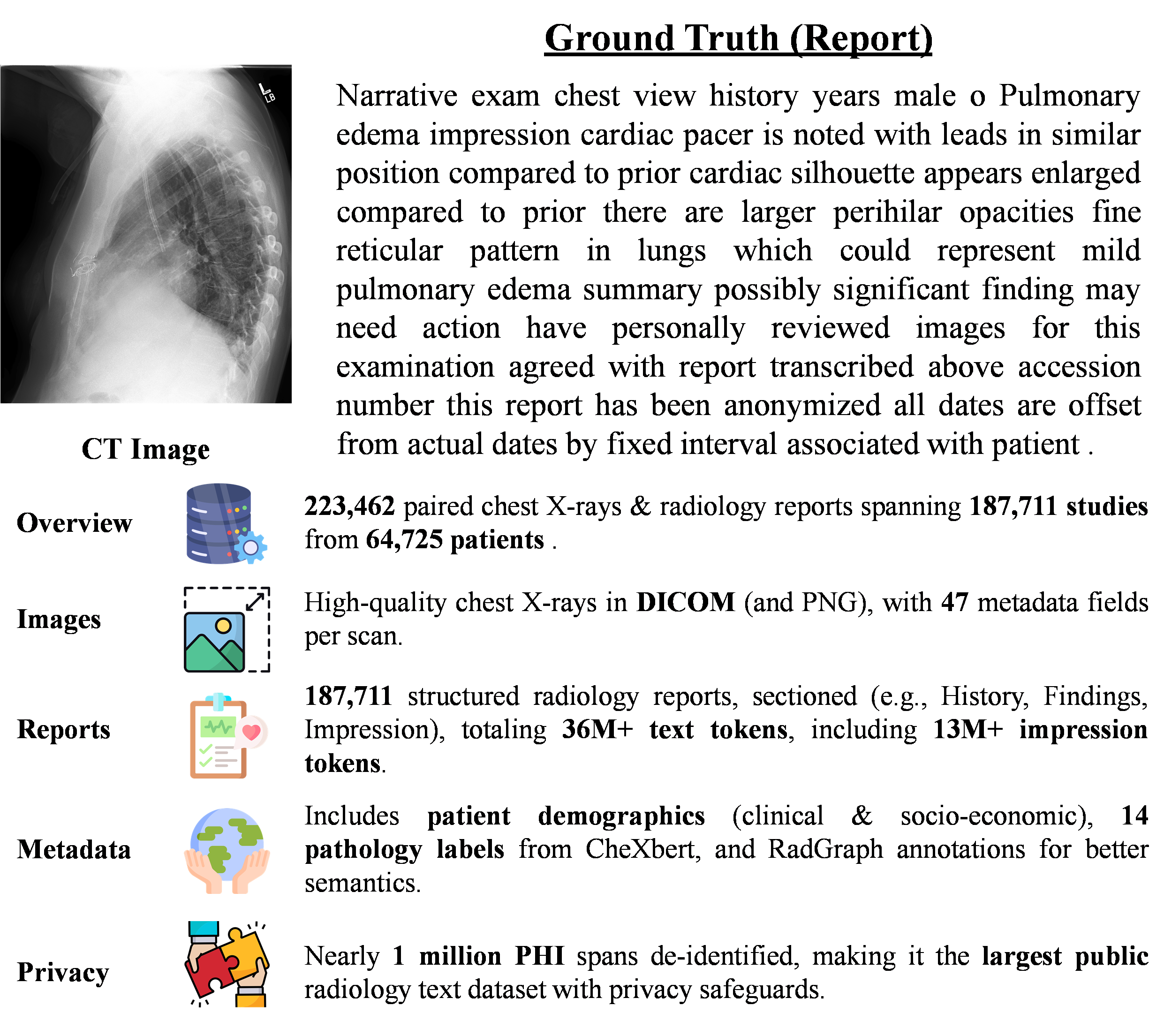}
\caption{Overview of CheXpert Plus dataset}
\label{fig6}
\end{figure}

We evaluate our method on CheXpert Plus~\cite{ref9}, a large-scale radiology dataset comprising 223,462 chest X-ray image–report pairs from 187,711 studies across 64,725 patients. Reports are de-identified and structured into 11 clinical sections, with 14 expert-defined pathology labels following the CheXpert schema. We follow the official protocol and adopt an 80/20 train–test split, ensuring no patient overlap between splits (Figure~\ref{fig6}).

We further evaluate on the IU-XRay dataset~\cite{ref10}, which contains 7,470 chest X-ray images and 3,955 reports, each organized into sections such as Comparison, Indication, Findings, and Impression. Following standard practice, we use predefined training, validation, and test splits, and focus evaluation on generating accurate and coherent Findings and Impression sections.

\subsection{Base Architecture}
\begin{figure}[h]
\centering
\includegraphics[width=1.1\textwidth]{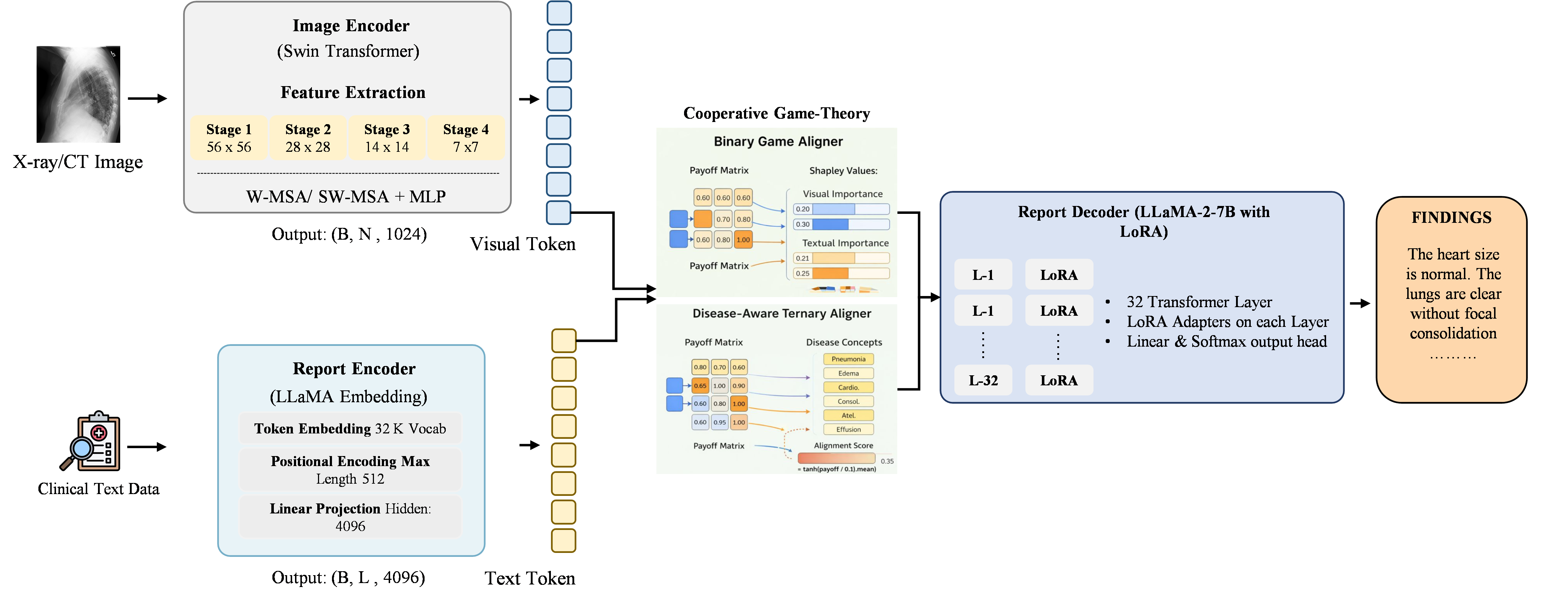}
\caption{The architecture of GTA-Net. We introduce a game theoretic alignment framework that models region level and disease level correspondence through binary and ternary games. For clarity, the figure illustrates a single image–report pair and a set of disease concepts used for disease level alignment.}
\label{fig1}
\end{figure}

The proposed GTA-Net framework comprises two core components: a visual encoder and a language model for radiology report generation (Figure~\ref{fig1}). The visual encoder extracts hierarchical representations from chest X-ray images, while the language model generates clinically coherent reports. Crucially, these components are tightly coupled through game-theoretic alignment modules that explicitly model interactions between visual regions, textual tokens, and disease concepts. This design enables GTA-Net to go beyond conventional encoder–decoder architectures by enforcing structured cross-modal alignment during generation.

\paragraph{Visual Encoder:} We adopt a Swin Transformer~\cite{ref11} as the visual backbone to extract high-dimensional representations from chest X-ray images. To enable parameter-efficient adaptation, we fine-tune the encoder using LoRA~\cite{ref12}, allowing the model to effectively leverage large-scale image data while maintaining computational efficiency.

\paragraph{Language Model:} For report generation, we employ a LLaMA-based causal language model~\cite{ref13}, adapted using LoRA for parameter-efficient fine-tuning. Visual features extracted by the Swin Transformer are projected into the language model’s input space and incorporated via prompt-based conditioning, enabling effective integration of visual context. This design guides the generation process to produce reports that are closely grounded in the underlying image content.

\subsection{Proposed Theoretic Alignment Predictor}
To capture both fine-grained and high-level alignments between medical images and generated reports, we introduce a GTA-Net that explicitly models region-level and disease-level correspondence. By treating visual regions, textual tokens, and disease concepts as interacting players, the framework quantifies their joint contributions to enforce consistent cross-modal grounding, leading to more accurate and clinically coherent report generation.

BinaryGameAligner (Region Word Alignment):  
The region–word alignment module models image regions and textual tokens as cooperative players in a binary game. Visual features (1024-dimensional) and text embeddings (4096-dimensional) are projected into a shared 512-dimensional latent space through learned linear transformations, enabling direct cross-modal interaction and alignment.
The payoff matrix $P \in \mathbb{R}^{B \times N_v \times N_t}$ quantifies cross-modal interactions between visual and textual elements, where $B$ denotes the batch size, $N_v$ the number of visual regions, and $N_t$ the number of text tokens. Each entry in $P$ is computed as the cosine similarity between normalized projected visual and textual feature embeddings, capturing the affinity of region--word pairs.

To estimate the contribution of each region--word alignment, we compute \textit{Shapley-inspired importance weights} by applying a softmax normalization over the aggregated payoffs, providing a principled and efficient approximation of each element's contribution to the overall alignment.
\[
v_i = \mathrm{softmax}\big(\mathrm{mean}_t(P_{i,:})\big)
\]
for visual regions and
\[
t_j = \mathrm{softmax}\big(\mathrm{mean}_v(P_{:,j})\big)
\]

DiseaseAwareTernaryAligner (Disease Level Alignment):
To ensure clinically coherent report generation, we introduce a Disease-Aware Ternary Aligner that models joint interactions among image features, report tokens, and structured disease concepts. Specifically, high-dimensional visual features (1024-d) and textual embeddings (4096-d) are projected into a shared 256-dimensional disease concept space via learned linear mappings, enabling consistent cross-modal reasoning at the clinical level.

We maintain a learnable embedding matrix for 14-disease concepts covering common chest pathologies. To stabilize training, all projections are $\ell_2$-normalized before computing compatibility scores via matrix multiplication. The disease compatibility scores are clamped to $[-1, 1]$ to prevent saturation. 

To model higher-order clinical consistency, we construct a ternary payoff matrix 
$T \in \mathbb{R}^{B \times N_d \times N_d}$, where $B$ denotes the batch size 
and $N_d$ the number of disease concepts. Each entry of $T$ captures the interaction 
between visual disease scores and textual disease scores, computed via their outer 
product to reflect joint cross-modal compatibility. 

We then aggregate these interactions into a scalar alignment score by averaging over 
disease dimensions and applying a temperature-scaled nonlinearity:
\[
s = \tanh\!\left(\frac{\mathrm{mean}(T)}{\tau}\right),
\]
where $\mathrm{mean}(\cdot)$ denotes averaging over all elements of $T$, and 
$\tau = 0.1$ is a temperature parameter that stabilizes optimization by smoothing 
gradients and bounding the magnitude of the alignment signal.
\subsection{Overview of Proposed GTA-Net Predictor}
Our model jointly optimizes multiple objectives during training. The primary language modeling loss $\mathcal{L}_{\text{LM}}$ is the standard cross-entropy loss for next token prediction. 

The binary alignment loss encourages region to word correspondence through contrastive learning. Given the payoff matrix $\mathbf{P}$ from BinaryGameAligner, we compute bidirectional cross entropy losses where positive pairs correspond to matched image-report pairs: $\mathcal{L}_{\text{binary}} = (\mathcal{L}_{\text{i2t}} + \mathcal{L}_{\text{t2i}}) / 2$. The ternary alignment loss $\mathcal{L}_{\text{ternary}} = -s$ maximizes the disease level alignment score, promoting semantic consistency at the pathology concept level.

To promote global cross-modal consistency, we introduce a cosine similarity regularization term that aligns aggregated visual and textual representations. Specifically, visual features $\mathbf{f}_v$ and textual features $\mathbf{f}_t$, obtained by averaging over spatial and sequence dimensions respectively, are projected into a shared 512-dimensional space via learnable linear mappings. The regularization objective is defined as:
\begin{equation}
\mathcal{L}_{\mathrm{reg}} = 1 - \mathbb{E}[\cos(\mathbf{f}_v, \mathbf{f}_t)]
\end{equation}

The overall alignment loss combines binary region--token alignment $\mathcal{L}_{\mathrm{binary}}$, ternary image--text--disease alignment $\mathcal{L}_{\mathrm{ternary}}$, and the global regularization term:
\begin{equation}
\mathcal{L}_{\mathrm{align}} = 0.7\,\mathcal{L}_{\mathrm{binary}} + 0.2\,\mathcal{L}_{\mathrm{ternary}} + 0.1\,\mathcal{L}_{\mathrm{reg}}
\end{equation}

The weighting coefficients (0.7, 0.2, 0.1) were selected based on preliminary experiments under different configurations, where this combination consistently yielded the best trade-off between fine-grained alignment, semantic consistency, and overall generation performance.

The final training objective integrates language modeling and alignment:
\begin{equation}
\mathcal{L}_{\mathrm{total}} = \mathcal{L}_{\mathrm{LM}} + \lambda\,\mathcal{L}_{\mathrm{align}}
\end{equation}
where $\mathcal{L}_{\mathrm{LM}}$ denotes the language modeling loss and $\lambda = 0.1$ is a weighting factor balancing generation and alignment objectives. Alignment losses are applied only during training; at inference, report generation relies solely on the language model. To improve computational efficiency, visual features used for alignment are extracted without gradient backpropagation.

\textbf{Stability Measures:} To ensure stable optimization, we implement several design choices. All feature projections use L2 normalization before similarity computation. Disease compatibility scores are clamped to $[-1, 1]$, and the ternary payoff is scaled by temperature $\tau = 0.1$ before applying tanh activation. These measures prevent training instabilities and improve convergence.

\section{Experimental Setting and Result}
\label{sec:experiments}
\subsection{Training GTA-Net framework procedure}
All experiments are implemented in PyTorch Lightning and trained end-to-end on NVIDIA H100/H200 GPUs (80GB) using bfloat16 mixed precision. Training requires approximately 23--25 hours. We optimize the model using AdamW with a learning rate of $1 \times 10^{-4}$, a batch size of 16, and 40 training epochs. The visual encoder is a Swin-Base Transformer (Swin-B), and the language backbone is LLaMA-2-7B, both adapted via LoRA for parameter-efficient fine-tuning. The training objective combines language modeling with alignment losses, where the alignment weight is set to $\lambda = 0.1$. Within the alignment objective, binary, ternary, and cosine regularization components are weighted at 0.7, 0.2, and 0.1, respectively. We evaluate performance using standard text generation metrics, including BLEU~\cite{ref14}, ROUGE~\cite{ref15}, and CIDEr~\cite{ref16}, which are widely used in medical report generation. All hyperparameters are selected based on validation performance.

\subsection{Quantitative Results}
Table \ref{tab:comparison_results} summarizes the quantitative results on the CheXpert Plus dataset. GTA-Net consistently outperforms all competing methods across all evaluation metrics. Compared to recent approaches such as R2GenKG~\cite{ref17} and AM-MRG~\cite{ref18}, GTA-Net achieves higher BLEU scores (B-1: 0.423, B-4: 0.221), indicating improved n-gram precision at both coarse and fine levels. More notably, it delivers substantial gains in ROUGE-L (0.421) and CIDEr (0.351), reflecting enhanced long-range coherence and stronger alignment with clinically relevant reference reports. On the IU-XRay dataset, GTA-Net maintains state-of-the-art or highly competitive performance. It surpasses prior methods, including AM-MRG~\cite{ref18} and HSA~\cite{ref19}, across all BLEU metrics (e.g., B-4: 0.321) and achieves the highest ROUGE-L score (0.423), demonstrating improved linguistic consistency and structural completeness. Although HSA~\cite{ref19}, and R2GenKG~\cite{ref23} report slightly higher CIDEr scores, GTA-Net exhibits more balanced performance across metrics, suggesting greater overall stability and semantic fidelity in generated reports.


\begin{table}[h!]
\caption{Performance comparison of GTA-Net model with existing models}
\label{tab:comparison_results}
\centering
\small
\begin{tabular}{l | p{3cm} | p{3cm} | c c c c c c}
\hline
\textbf{Dataset} & \textbf{Method} & \textbf{Publication} & \textbf{B-1} & \textbf{B-2} & \textbf{B-3} & \textbf{B-4} & \textbf{R-L} & \textbf{CIDEr} \\
\hline

\multirow{5}{*}{\textbf{CheXpert Plus}} 
& R2GenKG~\cite{ref17}   & arXiv 2025 & 0.376 & 0.234 & 0.155 & 0.106 & 0.269 & 0.125 \\
& AM-MRG~\cite{ref18}    & arXiv 2025 & 0.381 & 0.238 & 0.159 & 0.109 & 0.282 & 0.221 \\
& R2GenCSR~\cite{ref20}  & arXiv 2024 & - & - & - & 0.103 & 0.272 & 0.193 \\
& EMRRG~\cite{ref21}     & arXiv 2025 & 0.375 & 0.232 & 0.153 & 0.104 & 0.273 & 0.167 \\
& GTA-Net               & \textbf{Ours} & \textbf{0.423} & \textbf{0.329} & \textbf{0.304} & \textbf{0.221} & \textbf{0.421} & \textbf{0.351} \\

\hline

\multirow{8}{*}{\textbf{IU-XRay}} 
& KGAE~\cite{ref7}       & NeurIPS 2021 & 0.512 & 0.327 & 0.240 & 0.179 & 0.383 & - \\
& HRGR~\cite{ref8}       & NeurIPS 2018 & 0.438 & 0.298 & 0.208 & 0.151 & 0.322 & 0.343 \\
& R2GenKG~\cite{ref17}   & arXiv 2025 & 0.468 & 0.312 & 0.231 & 0.181 & 0.383 & \textbf{0.701} \\
& AM-MRG~\cite{ref18}    & arXiv 2025 & 0.489 & 0.339 & 0.253 & 0.192 & 0.384 & 0.613 \\
& HSA~\cite{ref19}       & MICCAI 2024 & 0.527 & 0.361 & 0.268 & 0.196 & 0.405 & 0.598 \\
& R2Gen~\cite{ref22}     & EMNLP 2020 & 0.470 & 0.304 & 0.219 & 0.165 & 0.371 & - \\
& R2GenGPT~\cite{ref23}  & Meta Radiology 2023 & 0.488 & 0.316 & 0.228 & 0.173 & 0.377 & 0.438 \\
& GTA-Net               & \textbf{Ours} & \textbf{0.530} & \textbf{0.378} & \textbf{0.323} & \textbf{0.321} & \textbf{0.423} & 0.423 \\

\hline
\end{tabular}
\end{table}
We compare GTA-Net with existing report generation methods, including R2GenKG \cite{ref17}, AM-MRG \cite{ref18}, and EMRRG \cite{ref21}. GTA-Net achieves the best performance (Table~\ref{tab:ce_metrics_chexpert}), with a Precision of 0.421, Recall of 0.357, and F1-score of 0.386. These gains are primarily attributed to the proposed cooperative game-theoretic alignment mechanism, which enables more explicit and fine-grained interaction between visual regions and clinical text than conventional cross-modal approaches.
\begin{table}[h!]
\centering
\caption{Quantitative comparison of clinical efficacy metrics on the CheXpert Plus dataset.}
\label{tab:ce_metrics_chexpert}
\small
\begin{tabular}{lcccc}
\hline
\textbf{Method} & \textbf{Publication} & \textbf{Precision} & \textbf{Recall} & \textbf{F1} \\
\hline
R2GenKG~\cite{ref17} & arXiv 2025 & 0.292 & 0.338 & 0.275 \\
AM-MRG~\cite{ref18} & arXiv 2025 & 0.396 & 0.318 & 0.336 \\
EMRRG~\cite{ref21} & arXiv 2025 & 0.341 & 0.273 & 0.272 \\
\hline
\textbf{GTA-Net} & \textbf{Ours} & \textbf{0.421} & \textbf{0.357} & \textbf{0.386} \\
\hline
\end{tabular}
\end{table}

\subsection{Ablation study}
The ablation study evaluates the contribution of each component in the proposed GTA-Net framework (Table~\ref{tab:t2}). The vision–language backbone serves as the baseline, achieving Precision 0.352, Recall 0.296, and F1-score 0.312. Incorporating \cite{ref24} the Binary Game Aligner improves region–word correspondence, increasing performance to 0.381 / 0.321 / 0.337. Further adding the Disease-Aware Ternary Aligner enables joint reasoning over visual features, textual tokens, and disease concepts, yielding 0.404 / 0.344 / 0.361. Finally, introducing auxiliary alignment objectives regularizes training and enhances semantic grounding, resulting in the best performance for the full GTA-Net model, with Precision 0.421, Recall 0.357, and F1-score 0.386.
\begin{table}[h]
\centering
\caption{Component-wise ablation study of GTA-Net on the CheXpert Plus dataset}
\label{tab:t2}
\small
\begin{tabular}{lccc}
\hline
\textbf{Model Variant} & \textbf{Precision} & \textbf{Recall} & \textbf{F1-Score} \\ 
\hline
Baseline Vision-Language Backbone & 0.352 & 0.296 & 0.312 \\
+ Binary Game Aligner & 0.381 & 0.321 & 0.337 \\
+ Disease-Aware Ternary Aligner & 0.404 & 0.344 & 0.361 \\
+ Auxiliary Alignment Objectives & 0.413 & 0.351 & 0.372 \\
\textbf{GTA-Net (Full Model)} & \textbf{0.421} & \textbf{0.357} & \textbf{0.386} \\
\hline
\end{tabular}
\end{table}
\subsection{Clinical Explainability Analysis}
To evaluate model interpretability, we conduct explainability analyses to assess the alignment between generated reports, visual evidence, and disease semantics. These visualizations provide qualitative evidence of whether the model attends to clinically relevant areas and produces semantically consistent descriptions.

\begin{figure}[h]
\centering
\includegraphics[width=0.99\textwidth]{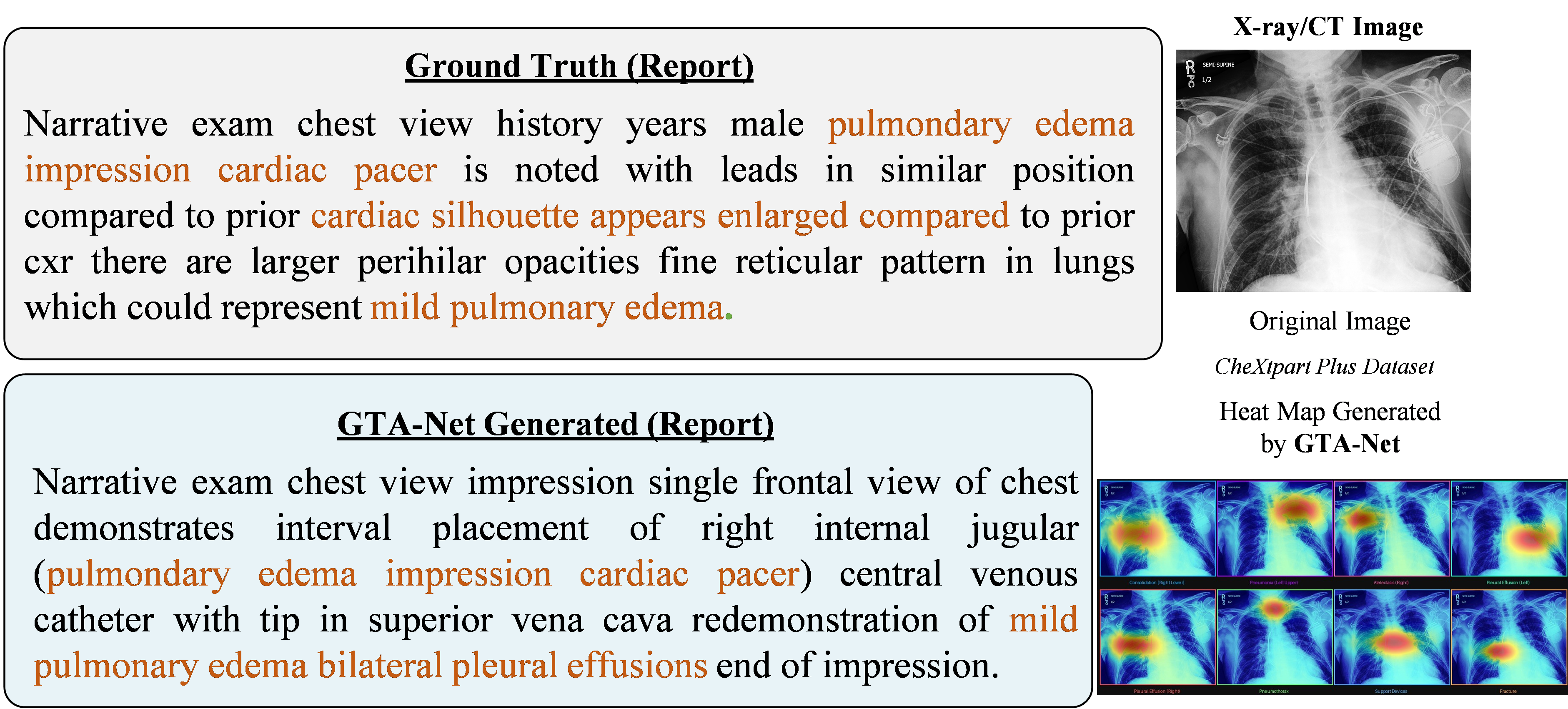}
\caption{Comparison of Ground Truth Report, Generated Report and XAI Highlighted Regions. The regions attended to by the model (e.g., pneumonia, mass and infiltration) are visualized using Grad CAM heatmaps overlaid on the original X-ray image.}
\label{fig11}
\end{figure}

Figure~\ref{fig11} presents qualitative results demonstrating the performance and interpretability of GTA-Net in chest X-ray report generation. The figure compares ground-truth and generated reports with color-coded annotations highlighting semantic alignment across clinical entities, including pathologies, anatomical structures, and imaging descriptors. GTA-Net accurately captures key diagnostic findings with strong consistency between visual evidence and textual descriptions. Attention heatmaps further illustrate that the model focuses on anatomically relevant regions, such as the lung fields and mediastinum, when predicting conditions like pleural effusion and pulmonary edema. These results indicate that GTA-Net achieves reliable cross-modal grounding while providing interpretable visual explanations, supporting both the clinical coherence of generated reports and transparency in decision-making.


\subsection{Discussion}
Our results demonstrate that explicitly modeling vision–language alignment as a cooperative game substantially improves radiology report generation, particularly in achieving semantic consistency across multiple levels of abstraction. The proposed Binary Game Aligner enforces fine-grained region–word correspondence through contrastive learning, enabling more precise grounding between visual findings and textual descriptions. Complementarily, the Disease-Aware Ternary Aligner enhances disease-level coherence by incorporating structured pathology representations, leading to more clinically faithful and context-aware reports. These components work collaboratively to deliver consistent gains across standard evaluation metrics, while also improving the quality of clinically meaningful descriptions rather than generating generic or templated text. This highlights the importance of explicit alignment objectives in promoting faithful and informative report generation. Moreover, GTA-Net establishes a principled and effective framework for cross-modal alignment, demonstrating that game-theoretic modeling can significantly enhance both the reliability and interpretability of medical vision–language systems.
\subsection{Limitations and Future work}
\subsubsection*{Limitations of GTA-Net}
\label{sec:limitations}
GTA-Net achieves strong empirical performance by introducing explicit game-theoretic alignment for cross-modal grounding; however, several aspects warrant further investigation. The cooperative alignment mechanism, based on binary and ternary payoff computations, introduces some additional training overhead, reflecting a trade-off between computational efficiency and richer semantic interaction modeling. In the current design, visual features used for alignment are extracted without gradient propagation, which stabilizes training but limits deeper end-to-end optimization between the visual encoder and alignment objectives. Furthermore, the Disease-Aware Ternary Aligner operates on a predefined set of 14-chest pathology concepts, enabling targeted clinical consistency, while future work could extend this formulation to a more scalable and dynamically learned concept space. Although GTA-Net demonstrates consistent improvements across standard evaluation metrics on CheXpertPlus and IU-XRay, broader validation including expert radiologist assessment and multi-modality datasets would further establish its clinical robustness and generalizability.
\subsubsection*{Future works}
\label{sec:future}
Despite strong empirical performance and consistent improvements over prior methods, GTA-Net has several limitations that suggest directions for future research.
\begin{itemize}
    \item While our cooperative alignment modules provide explicit and interpretable cross-modal grounding, they are trained on top of frozen visual features, limiting full end-to-end optimization between visual representation learning and alignment objectives. Enabling tighter integration could further enhance performance.
    \item The proposed disease-aware ternary alignment relies on a predefined set of clinical concepts, which may restrict coverage of rare or complex pathologies. Expanding this concept space or learning it dynamically could improve scalability and robustness.
    \item  Although GTA-Net achieves state-of-the-art results on standard benchmarks and improves clinical consistency, current evaluation protocols remain imperfect proxies for real-world reliability. Incorporating expert radiologist assessment and more clinically grounded metrics would provide a more comprehensive evaluation of model performance.
    \item Current experiments focus on chest X-ray datasets. Extending GTA-Net to other imaging modalities (e.g., CT and MRI) and larger, more diverse clinical populations would further validate its generalizability.
\end{itemize}

\section{Conclusion}
We introduced GTA-Net, a cooperative game-theoretic framework for automated chest X-ray report generation that formulates vision–language alignment as an explicit optimization objective. By modeling visual regions, textual tokens, and disease concepts as cooperative agents, GTA-Net enables joint learning of fine-grained region–word alignment and disease-level semantic consistency without requiring manual supervision. Extensive experiments demonstrate that GTA-Net consistently outperforms prior methods, achieving BLEU-4 scores of 0.221 on CheXpert Plus and 0.321 on IU-XRay, along with CIDEr scores of 0.351 and 0.423, respectively. Beyond quantitative gains, qualitative analyses show that the proposed alignment mechanism provides more interpretable and clinically grounded report generation. Our results establish cooperative game-theoretic alignment as an effective and principled approach for improving both the accuracy and reliability of medical vision–language models.
\section*{Declarations}
\begin{itemize}
\item \textbf{Ethics approval and consent to participate.}
\item \textbf{Funding.} No funding
\item \textbf{Declaration of competing interest.} The authors declare that they have no known competing financial interests or personal relationships that could have appeared to influence the work reported in this paper.
\item \textbf{Consent for publication.} Not applicable
\item \textbf{Data availability.} Data available at: \href{https://anonymous.4open.science/r/R2-09A3/}{Dataset}. 
\item \textbf{CRediT authorship contribution statement.} Saif Ur Rehman Khan, Imad Ahmed Waqar \& Muhammed Nabeel Asim: Conceptualization, Data curation, Methodology, Software, Validation, Writing original draft \& Formal analysis.Sebastian Vollmer \& Andreas Dengel: Conceptualization, Funding acquisition, Review. 
\end{itemize}

\end{document}